\def\year{2020}\relax
\documentclass[letterpaper]{article} 
\usepackage{aaai20}  
\usepackage{times}  
\usepackage{helvet} 
\usepackage{courier}  
\usepackage[hyphens]{url}  
\usepackage{graphicx} 
\usepackage{graphicx}  
\usepackage{amsmath}
\usepackage{multirow}
\title{Text Perceptron: Towards End-to-End Arbitrary-Shaped Text Spotting}
\author{Liang Qiao\textsuperscript{1}
	    Sanli Tang\textsuperscript{1}
	    Zhanzhan Cheng\textsuperscript{21}\thanks{Corresponding author.}
	    Yunlu Xu\textsuperscript{1}
	    Yi Niu\textsuperscript{1}
	    Shiliang Pu\textsuperscript{1}
	    Fei Wu\textsuperscript{2}\\
		\textsuperscript{1}Hikvision Research Institute, China; ~~~\textsuperscript{2}Zhejiang University, China\\
\{qiaoliang6,~~tangsanli,~~chengzhanzhan,~~xuyunlu,~~niuyi,~~pushiliang\}@hikvision.com~~~~~wufei@cs.zju.edu.cn
}
 
\begin{document}

\maketitle

\begin{abstract}
Many approaches have recently been proposed to detect irregular scene text and achieved promising results.
However, their localization results may not well satisfy the following text recognition part mainly because of two reasons:
1) recognizing arbitrary shaped text is still a challenging task, and 2) prevalent non-trainable pipeline strategies between text detection and text recognition will lead to suboptimal performances.
To handle this \emph{incompatibility} problem, in this paper we propose an end-to-end trainable text spotting approach named Text Perceptron.
Concretely, Text Perceptron first employs an efficient segmentation-based text detector that learns the latent text reading order and boundary information.
Then a novel Shape Transform Module (\emph{abbr.} STM) is designed to transform the detected feature regions into regular morphologies without extra parameters.
It unites text detection and the following recognition part into a whole framework, and helps the whole network achieve global optimization.
Experiments show that our method achieves competitive performance on two standard text benchmarks, i.e., ICDAR 2013 and ICDAR 2015, and also obviously outperforms existing methods on irregular text benchmarks SCUT-CTW1500 and Total-Text.
\end{abstract}

\section{Introduction}
Spotting scene text is a hot research topic due to its various applications such as invoice recognition and road sign reading in advanced driver assistance systems.
With the advances of deep learning, many deep neural-network-based methods  \cite{wang2012end,jaderberg2014deep,li2017towards,liu2018fots,he2018end} have been proposed for spotting text from a natural image, and have achieved promising results.
\begin{figure}[ht]
\centering
\includegraphics[width=1.0\columnwidth]{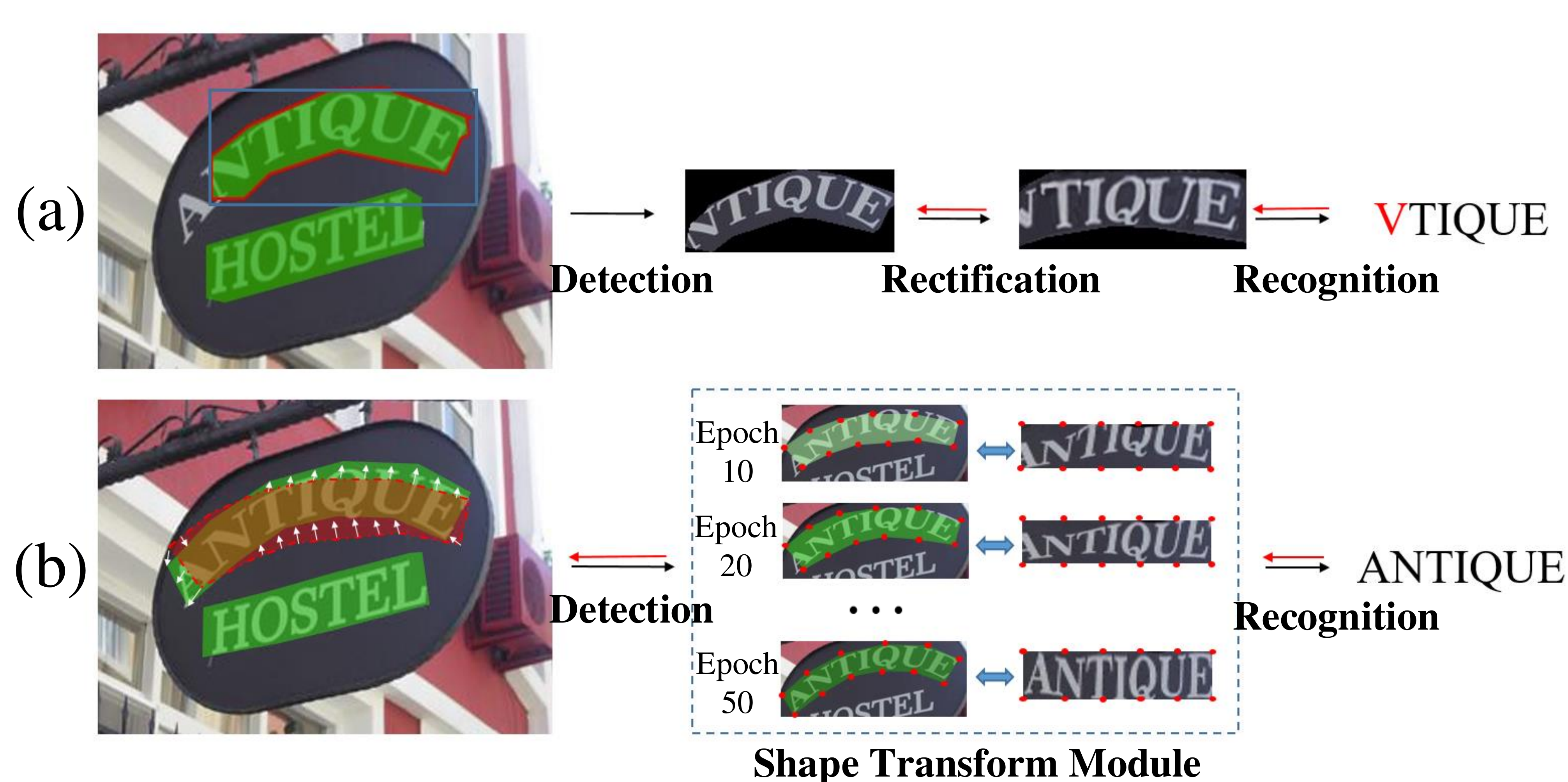}\\
   \caption{
   Illustration of the traditional pipelined text spotting process and Text Perceptron.
   Sub-figure (a) is a traditional pipeline strategy by combining text detection, rectification and recognition into a framework.
   Sub-figure (b) is an end-to-end trainable text spotting approach by applying the proposed STM.
   The black and red arrows mean the forward and backward processing, respectively.
   The red points denote generated fiducial points generated.
   }
\label{fig:0}
\end{figure}

However, in the real-world, many texts appear in arbitrary layouts (\emph{e.g.} multi-oriented or curved), which make quadrangle-based methods \cite{liao2017textboxes,zhou2017east,zhang2018feature} cannot be well adapted in many situations.
Some works \cite{dai2018fused,long2018textsnake,xie2018scene} began to focus on irregular text localization by segmenting text masks as detection results and achieved relatively good performance in terms of Intersection-over-Union (IoU) evaluation.
However, they still leave many challenges to the following recognizing task.
For example, a common pipeline of text spotting is to crop the masked texts within bounding-box regions, and then adopt a recognition model with rectification functions to generate final character sequences.
Unfortunately, such strategy decreases the robustness of text spotting mainly in two aspects:
1) one needs to design extra rectification network, like methods in \cite{luo2019moran} and \cite{zhan2019esir}, to transform irregular texts into regular ones.
In practice, it is hard to be optimized without human-labeled geometric ground truth, and also introduces extra computational cost.
2) Pipelined text spotting methods are not end-to-end trainable and result in suboptimal performance because the errors from the recognition model cannot be utilized for optimizing the text detector.
In Figure  \ref{fig:0}(a), although the text detector provides true positive results, the clipped text masks still lead to wrong recognition results.
We denote above problem \emph{incompatibility} between text detection and recognition.

Recently, two methods were proposed for spotting irregular text in the end-to-end manners.
\cite{lyu2018mask} proposed an end-to-end trainable network inspired by Mask-RCNN \cite{he2017mask}, aiming at reading irregular text character-by-character.
However, this approach loses the context information among characters, and also requires amounts of expenditure on character-level annotations.
\cite{sun2018textnet} attempted to transform irregular text with a perspective ROI module, but this operation has difficulty in handling some complicated distortions such as curved shapes.

These limitations motivate us to explore new and more effective method to spot irregular scene text.
Inspired by \cite{shi2016robust}, thin-plate splines (\emph{abbr.} TPS) \cite{bookstein1989principal} may be a feasible approach to rectify various-shaped text into regular form using a group of fiducial points. Although these points can be implicitly learned from cropped rectangular text by a deep spatial transform network \cite{jaderberg2015spatial}, the learning process of fiducial points is hard to be optimized. As a result, such methods are not robust especially for texts in some complex distortions.

In a more achievable way, we attempt to solve this problem as follows: 1) explicitly finding out a group of reliable fiducial points over text regions so that irregular text can be directly rectified by TPS, and 2) dynamically tuning fiducial points by back-propagating errors from recognition to detection.
Specifically, we develop a Shape Transform Module (\emph{abbr.} STM) to build a robust irregular text spotter and eliminate the \emph{incompatibility} problem.
STM integrates irregular text detection and recognition into an end-to-end trainable model, and iteratively adjusts fiducial points to satisfy the following recognition module.
As shown in Figure \ref{fig:0}(b), in the early training stage, despite high IoU in detection evaluation, the transformed text regions may not satisfy the recognition module.
With end-to-end training, fiducial points will be gradually adjusted to obtain better recognition results.

In this paper, we propose an end-to-end trainable irregular text spotter named Text Perceptron which consists of three parts:
1) A segmentation-based detection module which orderly describes a text region as four subregions: the \emph{center} region, \emph{head, tail} and \emph{top\&bottom} \emph{boundary} regions, detailed in Section 3. Here, boundary information not only helps separate text regions that are very close to each other, but also contributes to capture latent reading-orders.
2) STM for iteratively generating potential fiducial points and dynamically tuning their positions, which alleviates \emph{incompatibility} between text detection and recognition.
3) A sequence-based recognition module for generating final character sequences.

Major contributions of this paper are listed as follows:
1) We design an efficient order-aware text detector to extract arbitrary-shaped text.
2) We develop the differentiable STM devoting to optimizing both detection and recognition in an end-to-end trainable manner.
3) Extensive experiments show that our method achieves competitive results on two regular text benchmarks, and also significantly surpasses previous methods on two irregular text benchmarks.

\section{Related Works}
Here, we briefly review the recent advances in text detection and end-to-end text spotting.
\subsection{Text Detection}
Methods of text detection can usually be divided into two categories: anchor-based methods and segmentation-based methods.

\emph{Anchor-based methods}. These methods usually follows the technique of Faster R-CNN \cite{ren2015faster} or SSD \cite{liu2016ssd} that uses anchors to provide rectangular region proposals.
To overcome the significantly varying aspect ratios of texts, \cite{liao2017textboxes} designed long default boxes and filters to enhance text detection, and then \cite{liao2018textboxess} extended this work by generating quadrilateral boxes to fit the texts with perspective distortions.
\cite{ma2018arbitrary} proposed a rotated regional proposal network to enhance multi-oriented text detection.
To detect arbitrary-shaped text, many Mask RCNN \cite{he2017mask}-based methods, \emph{e.g.}, CSE \cite{liu2019Towards}, LOMO \cite{Zhang2019look} and SPCNet \cite{xie2018scene}, were developed to capture irregular texts and achieved good performance.

\emph{Segmentation-based methods}.
These methods usually learn a global semantic segmentation without region proposals, which is more efficient compared to anchor-based methods. Segmentation can easily be used to describe text in arbitrary shapes but highly relies on complicated post-processes to separate different text instances.
To solve this problem, \cite{wu2017self} introduced boundary semantic segmentation to reduce the efforts in post-proposing.
EAST \cite{zhou2017east} learned a shrink text region and directly regressed the multi-oriented quadrilateral boxes from text pixels.
 \cite{long2018textsnake} designed a series of overlapping disks with different radii and orientations to describe arbitrary-shaped text regions.
 \cite{Wang2019Shape} proposed a method that first generates text region masks with various shrinkage ratios and then uses a progressive expansion algorithm to produce the final text region masks.
 \cite{xu2019textfield} predicted each text pixel and assigned them with a regression value denoting the direction to its nearest boundary to help separate different texts.

\subsection{Text Spotting}
Most of existing text-spotting methods \cite{liao2018textboxess,liao2017textboxes,wang2012end}  generally first localize each text with a trained detector such as \cite{zhou2017east} and then recognize the cropped text region with a sequence decoder \cite{shi2017end}.
For sufficiently exploiting the complementarity between detection and recognition, some works \cite{he2018end,li2017towards,liu2018fots} were proposed to jointly detect and recognize text instances in an end-to-end trainable manner, which utilized the recognition information to optimize the localization task.
However, these methods are incapable of spotting arbitrary-shaped text due to the irrationality of rectangles or quadrangles.
To address these problems, \cite{sun2018textnet} adopted a perspective ROI transforming module to rectify perspective text, but this operation still has difficulty in handling serious curved text.
\cite{lyu2018mask} proposed an end-to-end text spotter inspired by Mask-RCNN for detecting arbitrary-shaped text character-by-character, but this method loses the context information among characters and also requires character-level location annotations.

\section{Methodology}
\subsection{Overview}
\begin{figure*}[ht]
\centering
\includegraphics[width=0.95\textwidth]{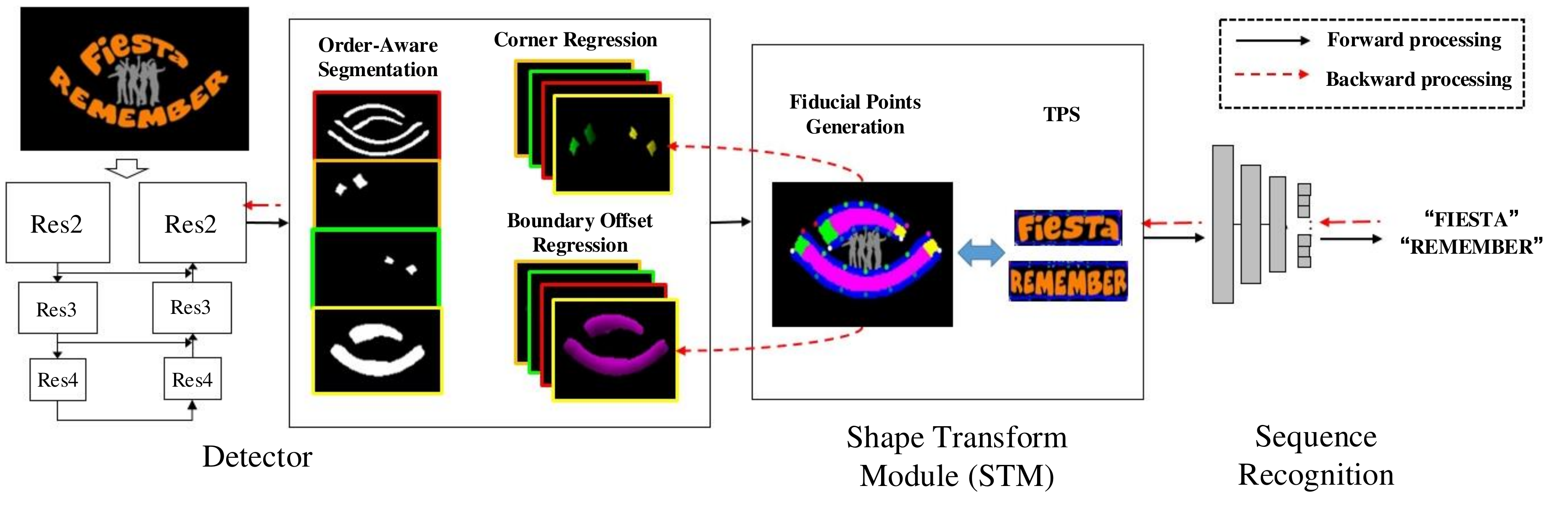}\\
\caption{The workflow of Text Perceptron. The black and red arrows separately mean the forward and backward process.}
\label{framework}
\end{figure*}

We propose a text spotter named Text Perceptron whose overall architecture is shown in Figure \ref{framework}, which consists of three parts:

(1) The text detector adopts ResNet \cite{he2016deep} and Feature Pyramid Network (\emph{abbr}. FPN) \cite{lin2017feature} as backbone, and is implemented by simultaneously learning three tasks: an order-aware multiple-class semantic segmentation, a corner regression, and a boundary offset regression. In this way, the text detector can localize arbitrary-shaped text and achieve state of the art on text detection.

(2) STM is responsible for uniting text detection and recognition into an end-to-end trainable framework.
This module iteratively generates fiducial points on text boundaries based on the predicted score and geometry maps, and then applies the differentiable TPS to rectify irregular text into regular form.

(3) The text recognizer is used to generate the predicted character sequences, which can be any traditional sequence-based method, such as CRNN \cite{shi2017end}, attention-based method \cite{cheng2017focus}.

\subsection{Text Detection Module}
\subsubsection{Order-aware Semantic Segmentation}
The text detector learns a global multi-class semantic segmentation, which is much more efficient than those Mask-RCNN-based methods.
Inspired by \cite{xue2018accurate}, we introduce text boundary segmentation to separate different text instances.
Considering text with arbitrary shapes, we further category boundaries into \emph{head}, \emph{tail}, and \emph{top\&bottom boundary} types, respectively.
In Figure \ref{labelgeneration}, the green, yellow, blue and pink regions separately denote the \emph{head}, \emph{tail}, \emph{top\&bottom} boundaries and the \emph{center} text region.
Here, \emph{head} and \emph{tail} also capture potential information about text reading order (\emph{e.g.} top to bottom for vertical text).
Therefore, we learn the text detector by conducting the multi-class semantic segmentation task using several binary Dices Coefficient Loss \cite{milletari2016v} (denoted by $\mathcal{L}_{cls}$).

\subsubsection{Corner and Boundary Regressions}
To boost the arbitrary-shaped segmentation performance as well as provide position information for fiducial points, we integrate two other regression tasks into the learning process, as shown in Figure \ref{labelgeneration} (c) and (d),
\begin{itemize}
\item \emph{Corner Regression.}
For pixels in \emph{head} and \emph{tail} regions, we regress the offsets (\emph{e.g.} the $\Delta dx_1, \Delta dy_1, \Delta dx_2$ and $\Delta dy_2$) to their corresponding two corner points, which is denoted by $\mathcal{L}_{corner}$.

\item \emph{Boundary Offset Regression.}
For pixels in \emph{center} region, we regress the vertical and horizontal offsets to their nearest boundaries (\emph{e.g.} the $\Delta dx_1^\prime, \Delta dy_1^\prime, \Delta dx_2^\prime$ and $\Delta dy_2^\prime$), which is denoted by $\mathcal{L}_{boundary}$.
\end{itemize}
Here, we adopt a proximity regression strategy to solve the inaccurate large-offset regression problem like in EAST \cite{zhou2017east}.
That is, the Corner Regressions only regress their neighboring corresponding corners.
In the Boundary Offset Regression, we can simply ignore or lower the loss weights of regression value generated from the larger side (\emph{e.g.} $\Delta dx_1^\prime, \Delta dx_2^\prime$ for a horizontal text). In this way, our detector can well describe the texts with very large width-height ratios.
Both of two regressions are trained with Smooth-L1 loss:
\begin{equation}
\mathcal{L}_{corner} ~or~ \mathcal{L}_{boundary} = \left\{
\begin{array}{lr}
0.5(\sigma z)^2       \qquad   |z| < 1/\sigma^{2} \\
 |z| - 0.5/\sigma^2   \quad    \text{otherwise} \\
\end{array}
\right. ,
\end{equation}
where $z$ is the geometry offset value, and $\sigma$ is a tunable parameter (default by 3).

\subsubsection{The Detection Inference}
In the forward process, we generate predicted segmentation maps by orderly overlaying the segmented \emph{center}, \emph{head}, \emph{tail}, and \emph{top\&bottom} boundary feature maps.
Subsequently, text instances can be found as connected-regions of \emph{center} pixels.
We see that all text instances are easily separated by boundaries, and different \emph{head} (or \emph{tail}) regions will also be separated by \emph{up\&bottom} boundary region. Therefore, each \emph{center} region can be matched with a neighboring pair of \emph{head} and \emph{tail} region during the pixel traversal process.
Specifically, for text with more than 1 \emph{head} (or \emph{tail}) regions, we choose the one with the maximum area as its \emph{head} (or \emph{tail}).
While for predicted \emph{center} text regions without corresponding \emph{head} or \emph{tail} region, we just treat them as false positives and filter them out.

\subsubsection{Ground-Truth Generation}
\begin{figure}
\centering
\includegraphics[width=0.9\columnwidth]{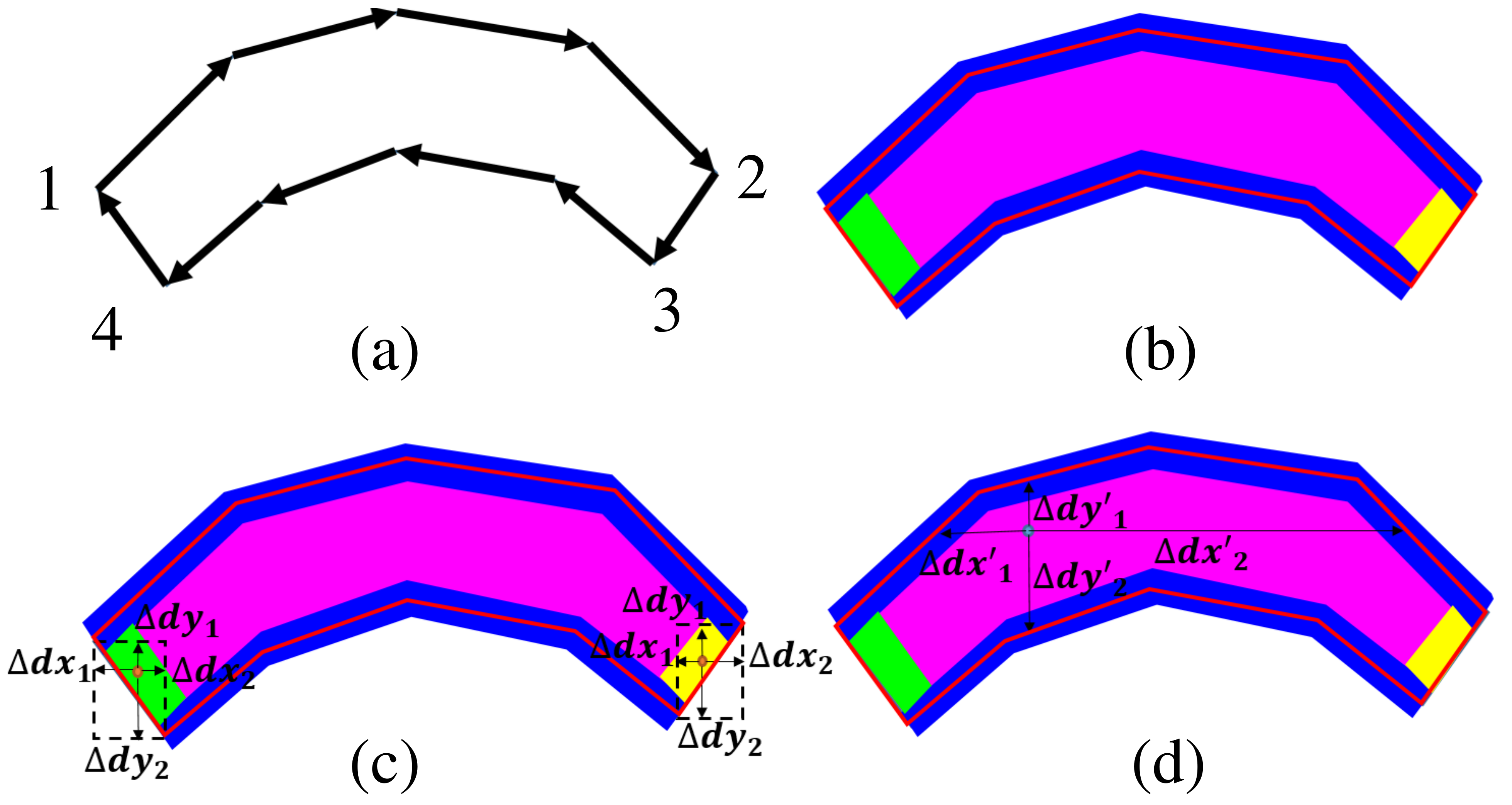}\\
\caption{The label generation process.}
\label{labelgeneration}
\end{figure}

The process of ground-truth of segmentation and geometry map can be divided into three steps, as shown in Figure \ref{labelgeneration}.

\emph{(1) Identifying four corners.}
We denote the 1st and 4th corners as the two corners in the \emph{head} region, while the 2nd and 3rd corners are corresponding to the \emph{tail} region, as shown in Figure \ref{labelgeneration}(a). This weak-supervised information is not provided by most of the datasets, but we found that in general, polygon points $\{P_1^\prime, ..., P_{M}^\prime\}$ are usually annotated from the left-top corner to the left-bottom corner in a clockwise manner for text instances.
Differently, for polygon annotations with a fixed number of points like SCUT-CTW1500 \cite{liu2019curved}, we can directly identify the four corner points by their indexes.
However, for annotations with varying number of points like Total-Text \cite{ch2017total}, we can only obtain the 1st corner ($P_1'$) and 4th corner ($P_{M}'$).
To search the 2nd and 3rd corners, we design a heuristic corner estimating strategy based on the assumptions that 1) two boundaries neighboring \emph{tail} are nearly parallel, and 2) two neighbor interior angles of \emph{tail} are closed to $\frac{\pi}{2}$.
Therefore, the probable 2nd corner can be estimated as:
\begin{equation}
\arg\min_{P_i^\prime} [\gamma(|\angle P_{i}^\prime - \frac{\pi}{2}|+|\angle P_{i+1}^\prime -\frac{\pi}{2}|)+ |\angle P_{i}^\prime + \angle P_{i+1}^\prime - \pi|]
\end{equation}
where $\angle P_{i}^\prime$ is the degree of interior angle for polygon point $P_i^\prime$, and $\gamma$ is a weighting parameter (default by 0.5).
Then the point $P_{i+1}^\prime$ following $P_i^\prime$ is treated as the 3-rd corner point.
Specifically, for vertical text annotated from the top-left corner, we reassign its top-right corner as the 1st key corner.

\emph{(2) Generating score maps.}
Figure \ref{labelgeneration}(b) shows the generated score maps. We firstly generate the \emph{center} text regions follows by their annotations and then generate boundaries by referring to the shrink and expansion mechanism used in \cite{wu2017self}.
Differently, the \emph{head} and \emph{tail} score maps are generated by only applying the shrink operation, which submerges part of the \emph{center} region.
And \emph{top\&bottom} boundary region is then generated by applying both the expansion and shrink operations, which will partly submerge all of the other regions.
In this way, we need less effort on post-processing to separate different text instances and it is easy to match their relative \emph{head} (or \emph{tail}) region with a \emph{center} region.
Boundary widths are constrained as $\delta$$\times$$minLen$, where $minLen$ is the minimum length of edges in the text polygon and $\delta$ is a ratio parameter.
Here, we set $\delta$$\text{=}$$0.2$ for \emph{top\&bottom} boundaries and $\delta$$\text{=}$$0.3$ for \emph{head} and \emph{tail}.

\emph{(3) Generating geometry maps.}
As mentioned in  Corner and Boundary Regression, pixels belonging to the \emph{head} region are assigned geometry offset values in 4 channels ($\Delta dx_1, \Delta dy_1, \Delta dx_2$ and $\Delta dy_2$) corresponding the 1st and 4th key corner, as shown in Figure \ref{labelgeneration}(c).
Similarly, the geometry map of the \emph{tail} region is also formed in 4 channels.
The geometry values of the \emph{center} text region are computed as the horizontal and vertical offsets to the nearest boundaries, shown as $\Delta dx'_1, \Delta dy'_1, \Delta dx'_2$ and $\Delta dy'_2$ in Figure \ref{labelgeneration}(d).

\subsection{Shape Transform Module}
STM is designed to iteratively generate initial fiducial points around text instances and transform text feature regions into regular shapes with the supervision of following recognition.
\subsubsection{Fiducial Points Generation}
With the learned segmentation maps and geometry maps, we propose to generate preset $2$$\times$$N$ potential fiducial points ($N$$\ge$$2$) for each text instance, denoted as $\{P_1,...,P_N, P_{N+1},...,P_{2\times N}\}$, which can be divided into two stages.
\begin{figure}
\centering
\includegraphics[width=0.9\columnwidth]{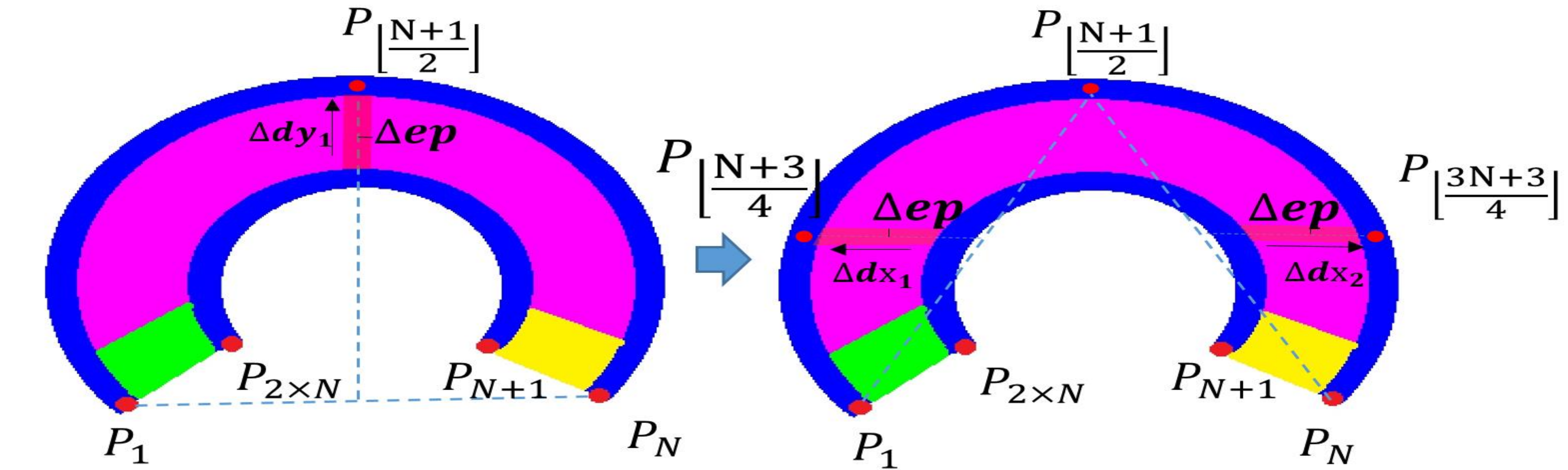}\\
\caption{The fiducial points generation process.}
\label{fig:fiducial-points}
\end{figure}

\emph{(1) Generating four corner points.}
We first obtain the positions of four corner fiducial points for each text feature region by averaging the coordinate of pixels with their predicted offsets in corresponding boundaries.
Taking the 1-st corner point ($P_1$) as an example, it is computed based on all pixels in the \emph{head} region $\mathcal{R}_\mathcal{H}$, and formalized by
\begin{equation}
P_1  = \left(\frac{\sum_{(x,y) \in \mathcal{R}_\mathcal{H}} (x + \Delta dx)}{||\mathcal{R}_\mathcal{H}||}, \frac{\sum_{(x,y) \in \mathcal{R}_\mathcal{H}} (y + \Delta dy)}{||\mathcal{R}_\mathcal{H}||}\right)
\end{equation}
where $||.||$ means the number of pixels in $\mathcal{R}_\mathcal{H}$, and $\Delta dx,\Delta dy$ mean the predicted corner offsets corresponding to ${P}_1$.
The other three corner points ($P_N$ in $\mathcal{R}_\mathcal{H}$, $P_{N+1}$, $P_{2\times N}$ in tail region $\mathcal{R}_\mathcal{T}$) can be calculated similarly.

\emph{(2) Generating other fiducial points.}
After obtaining four corner fiducial points, the other fiducial points can be located using a dichotomous method.
This strategy is suitable for any arbitrary shaped text even serious curved or in different reading orders.

An example of the generation process is shown in Figure \ref{fig:fiducial-points}.
We firstly connect $P_1$ and $P_N$, and judge whether the connected line has a longer span in horizontal direction or vertical direction. Without loss generality, if it has a longer span in horizontal direction as shown, we calculate a middle point $P_{\lfloor(N+1)/2\rfloor}$  between $P_1$ and $P_N$ whose x-coordinate formed as:
\begin{equation}
x_{mid}=\frac{\lceil(N-1)/2\rceil}{N-1}\times P_{1,x} + \frac{\lfloor(N-1)/2\rfloor}{N-1}\times P_{N,x}
\end{equation}
Then we use the learned boundary offsets from detector to predict the y-coordinate of $P_{\lfloor(N+1)/2\rfloor}$. Concretely, we define the \emph{band} region $\mathcal{B}_i$ as the part of the \emph{center} region $\mathcal{R}_C$:
\begin{equation}
\mathcal{B}_{\lfloor\frac{1+N}{2}\rfloor}=\{(x,y)\in\mathcal{R}_\mathcal{C}|x\in[x_{mid}-\Delta ep,x_{mid}+\Delta ep]\}
\end{equation}
where $\Delta ep$ defines the range of the band region (default by 3). Similar to the generation of four corner fiducial points, we can use all pixels in the corresponding band region to predict an average y-coordinate for this fiducial point. Then, the coordinate of $P_{\lfloor(N+1)/2\rfloor}$ can be formed as:
\begin{equation}
P_{\lfloor\frac{1+N}{2}\rfloor}=\left(x_{mid}, \frac{\sum_{(x_t,y_t)\in\mathcal{B}_{\lfloor\frac{1+N}{2}\rfloor}} y_t + \Delta dy'_t}{||\mathcal{B}_{\lfloor\frac{1+N}{2}\rfloor}||}\right)
\end{equation}
where $\Delta dy'_t$ is the learned boundary offset value to the top-boundary ($\Delta dy'_1$).
This process can be iteratively conducted using corresponding $\Delta dx'_t$ or $\Delta dy'_t$ until all of the fiducial points be calculated.
Similarly, the fiducial points on the bottom boundary can be calculated by connecting $P_{N+1}$ and $P_{2\times N}$ and using the same strategy.

\subsubsection{Shape Transformation}
With the generated potential fiducial points on text boundaries, we can explicitly transform an irregular feature region $\mathcal{R}$ into a regular form $\mathcal{R}^*$.
Here, fiducial points are mapped into some preset positions of the transformed feature map by directly applying TPS to the original feature regions.
Specifically, we transform all feature regions into a region with width $W$ and height $H$:
\begin{equation}
\mathcal{R}^* = TPS^{-1}(\mathcal{P}, \mathcal{R}),
\end{equation}
where the fiducial point $P_i \in \mathcal{P}$ will be mapped into:
\begin{equation}
P^*_i = \left\{
\begin{array}{lr}
\left((i-1)\times\frac{H-2\times\Delta w}{N-1}+\Delta w, ~\Delta h  \right) ,  ~ 1 \le i \le N\\
\left((2\times N -i)\times\frac{H-2\times\Delta w}{N-1}+\Delta w, ~ H-\Delta h    \right), \\
~~~\qquad \qquad \qquad \qquad\qquad \qquad   N < i \le 2\times N\\
\end{array}
\right.
\end{equation}
where $\Delta w$ and $\Delta h$ are preset offsets (default by 0.1$\times W$ and 0.1$\times H$) to preserve space for fiducial points tuning.

Then, all text feature regions are packed into a batch and sent to the following recognition part.
Here, we assume that the final predicted character strings $Y$ are generated as:
\begin{equation}
Y = Recog(\mathcal{R}^*),
\end{equation}
where `$Recog$' is the sequence recognition process.

\subsubsection{Dynamically Finetuning Fiducial Points}
The assumption here is that although text detector supervised by polygon annotations can generate satisfying polygon masks, the results may not always suitable for the following recognition.
To avoid the suboptimal problem and improve overall performance, Text Perceptron will back-propagate differences from `$Recog$' to each pixel value in $\mathcal{R}$ via STM, i.e.
\begin{equation}
\Delta \mathcal{R} = \frac{\partial Y}{\partial \mathcal{R}^*} \frac{\partial \mathcal{R}^*}{\partial \mathcal{R}}.
\end{equation}
Then we can calculate the adjustment values of $\mathcal{P}$ by
\begin{equation}
\Delta \mathcal{P} = \frac{\partial Y}{\partial \mathcal{R}^*} \frac{\partial \mathcal{R}^*}{\partial \mathcal{R}} \frac{\partial \mathcal{R}}{\partial \mathcal{P}}.
\end{equation}
Furthermore, we back-propagate $\Delta \mathcal{P}$ to the corresponding geometry maps in \emph{head}, \emph{tail} and \emph{band} regions.
Formally, for each pixel $pi$, we have
\begin{equation}
\Delta pi = \Delta \hat{pi} + \frac{\Delta \mathcal{P}}{||\mathcal{R}_{\mathcal{R}^*}||},
\end{equation}
where $\mathcal{R}_{\mathcal{R}^*} \in \{\mathcal{R}_{\mathcal{H}}, \mathcal{R}_{\mathcal{T}}, \mathcal{B}\}$ and $\Delta \hat{pi}$ is calculated from $\mathcal{L}_{corner}$ or $\mathcal{L}_{boundary}$.

\subsection{End-to-End Training}
Our recognition part can be implemented by any sequence-based recognition network, such as CRNN \cite{shi2017end} or \cite{cheng2017focus}.

The loss of the whole framework contains the following parts: the order-aware multi-class semantic segmentation, the corner regressions for pixels in \emph{head} and \emph{tail}, the boundary offset regression for pixels in the \emph{center} region and the word recognition, that is,
\begin{equation}
\mathcal{L} = \mathcal{L}_{cls} + \lambda_{b}\mathcal{L}_{corner}+\lambda_{c}\mathcal{L}_{boundary} + \lambda_{r}\mathcal{L}_{recog},
\end{equation}
where $\lambda_{b}$, $\lambda_{c}$ and $\lambda_{r}$ are auto-tunable parameters, and $\mathcal{L}_{recog}$ is the loss from recognition.

Since learning fiducial points highly depends on the segmentation map learning, we use a soft loss weight strategy to automatically tune $\lambda_{b}$, $\lambda_{c}$ and $\lambda_{r}$.
In other words, in the first few epochs, fiducial points are mainly adjusted by regression tasks; while at the last few epochs, points are mainly restricted to recognition.
Formally,
\begin{equation}
\lambda_{b} = \lambda_{c} = \lambda^* - \max{(0.02\times E,0.5)},
\end{equation}
\begin{equation}
\lambda_{r} =\min{( \max{(-0.1 + 0.02\times E, 0)}, \lambda_{r}^*)},
\end{equation}
where $E$ is the number of training epochs, and $\lambda^*$ and $\lambda_{r}^*$ separately control the maximum loss weight of regression and recognition.
In our experiments, we set $\lambda^*=0.6$ and $\lambda_{r}^*=0.8$.

\section{Experiments}

\begin{table*}[htbp]
\centering
\scalebox{0.87}{
\begin{tabular}{|c|l|c|c|c|c|c|c|c|c|c|c|}
\hline
\multirow{2}{*}{Dataset} & \multirow{2}{*}{Method} & \multicolumn{4}{c|}{Detection} & \multicolumn{3}{c|}{End-to-End} & \multicolumn{3}{c|}{Word Spotting} \\ \cline{3-12}
 & &P & R & F & FPS & S & W & G & S & W & G  \\
\hline
\multirow{10}{*}{IC13}
&Textboxes \shortcite{liao2017textboxes} & 88.0 & 83.0 & 85.0 & 1.37 & 91.6 & 89.7 & 83.9 & 93.9 & 92.0 & 85.9  \\
&Li et al. \shortcite{li2017towards}& 91.4 & 80.5 & 85.6 & - & 91.1 & 89.8 & 84.6 & 94.2 & 92.4 & 88.2 \\
&TextSpotter \shortcite{buvsta2017deep} &- & - & - & -  & 89.0 & 86.0 & 77.0 & 92.0 & 89.0 & 81.0 \\
&He et al. \shortcite{he2018end} & 91.0 & 88.0 & 90.0 & - & 91.0 & 89.0 & 86.0 & 93.0 & 92.0 & 87.0\\
&FOTS \shortcite{liu2018fots} & - & - & 88.2 & \textbf{23.9} & 88.8 & 87.1 & 80.8 & 92.7 & 90.7 & 83.5 \\
&TextNet* \shortcite{sun2018textnet}& 93.3 & \textbf{89.4} & 91.3 & - & 89.8 & 88.9 & 83.0 & 94.6 & \textbf{94.5} & 87.0 \\
&Mask TextSpotter* \shortcite{lyu2018mask} & \textbf{95.0} & 88.6 & \textbf{91.7} & 4.6 & \textbf{92.2} & \textbf{91.1} & \textbf{86.5} & 92.5 & 92.0 & {88.2}  \\
\cline{2-12}
&Ours (2-stage) & 92.7 & 88.7 & 90.7 & 10.3 & 90.8 & 90.0 & 84.4 & 93.7 & 93.1 & 86.2\\
&Ours (End-to-end) & 94.7 & 88.9 & \textbf{91.7} & 10.3 & 91.4 & 90.7 & 85.8 & \textbf{94.9} & 94.0 & \textbf{88.5}\\
\hline
\hline
\multirow{9}{*}{IC15}
&EAST \shortcite{zhou2017east} & 83.6 & 73.5 & 78.2 & \textbf{13.2} & - & - & - & - & - & - \\
&TextSnake* \shortcite{long2018textsnake} & 84.9 & 80.4 & 82.6 & 1.1  & - & - & - & - & - & -\\
&SPCNet* \shortcite{xie2018scene} & 88.7 & 85.8 & 87.2 & - & - & - & - & - & - & -\\
&PSENet-1s* \shortcite{Wang2019Shape} & 86.9 & 84.5 & 85.7 & 1.6  & - & - & - & - & - & -\\
&TextSpotter \shortcite{buvsta2017deep} & - & - & - & - & 54.0 & 51.0 & 47.0 & 58.0 & 53.0 & 51.0\\
&He et al. \shortcite{he2018end} & 87.0 & \textbf{86.0} & 87.0 & - & \textbf{82.0} &\textbf{ 77.0} & 63.0 & \textbf{85.0} & \textbf{80.0} & 65.0\\
&FOTS \shortcite{liu2018fots} & 91.0 & 85.2 & \textbf{88.0} & 7.8 & 81.1 & 75.9 & 60.8 & 84.7 & 79.3 & 63.3 \\
&TextNet* \shortcite{sun2018textnet} & 89.4 & 85.4 & 87.4 & -  & 78.7 & 74.9 & 60.5 & 82.4 & 78.4 & 62.4\\
&Mask TextSpotter* \shortcite{lyu2018mask} & 91.6 & 81.0 & 86.0 & 4.8   & 79.3 & 73.0 & 62.4 & 79.3 & 74.5 & 64.2 \\
\cline{2-12}
&Ours (2-stage)  & 91.6 & 81.8 & 86.4 & 8.8 & 78.2 & 74.5 & 63.0  & 80.6 &  76.6 & 65.5\\
&Ours (End-to-end) & \textbf{92.3} & 82.5 & 87.1 & 8.8 & 80.5 & 76.6 & \textbf{65.1} & 84.1 & 79.4 & \textbf{67.9}\\
\hline
\end{tabular}
}
\caption{Results on IC13 and IC15. `P', `R' and `F' separately mean the `Precision', `Recall' and `F-Measure'. `S', `W' and `G' mean recognition with strong, weak and generic lexicon, respectively. Superscript `*' means that the method considered the detection of irregular text.}
\label{tb:1}
\end{table*}

\subsection{Datasets}
The datasets used in this work are listed as follows:

\emph{SynthText 800k} \cite{gupta2016synthetic} contains 800k synthetic images that are generated by rendering synthetic text with natural images, and it is used as the pre-training dataset.

\emph{ICDAR2013} \cite{karatzas2013icdar} (\emph{abbr.} IC13) is collected as the focused scene text, which is mainly horizontal text containing 229 training images and 233 testing images.

\emph{ICDAR2015} \cite{karatzas2015icdar} (\emph{abbr.} IC15) is collected as incidental scene text consisting of many perspective text. It contains 1000 training and 500 testing images.

\emph{Total-Text} \cite{ch2017total} consists of multi-oriented and curve text and is therefore one of the important benchmarks in evaluating shape-robust text spotting tasks.
It contains 1255 training and 300 testing images, and each text is annotated by a word-level polygon with transcription.

\emph{SCUT-CTW1500} \cite{liu2019curved} (\emph{abbr.} CTW1500) is a curved text benchmark consists of 1000 training and 500 testing images.
In contrast to Total-Text, all text instances are annotated with 14-point polygons in the line-level.

\subsection{Implementation Details}

The detector uses ResNet-50 as the backbone and further be modified following the suggestions from \cite{huang2017speed} for obtaining dense features.
We remove the fifth stage, modify $conv4\_1$ layer with stride=1 instead of 2, and apply atrous convolution for all subsequent layers to maintain enough receptive field.
Training loss is calculated from the outputs of three stages: the fourth stage ($8$$\times$), the third stage ($8$$\times$), and the second stage ($4$$\times$) feature maps of FPN, and testing is only conducted on $4$$\times$ feature map.
We directly adopt the attention-based network described in \cite{cheng2017focus} as the recognition model.
All experiments are implemented in Caffe with 8 32GB-Tesla-V100 GPUs. The code will be published soon.

\textbf{Data augmentation.}
We conduct data augmentation by simultaneously 1) randomly scaling the longer side of input images with length in range of [$720$, $1600$],  2) randomly rotating the images with the degree in range of [$-15^\circ, 15^\circ]$, and 3) applying random brightness, jitters, and contrast on input images.

\textbf{Training details.}
The networks are trained by SGD with batch-size=8, momentum=0.9 and weight-decay=$5\times 10^{-4}$. For both detection and recognition part, we separately pre-train them on SynthText for 5 epochs with initial learning rate $2\times 10^{-3}$. Then, we jointly fine-tune the whole network using the soft loss weight strategy mention previously on each dataset for other 80 epochs. The initial learning rate is  $1\times 10^{-3}$. The learning rate will be divided by 10 for every 20 epochs.
Online hard example mining (OHEM) \cite{shrivastava2016training} strategy is also applied for balancing the foreground and background samples.

\textbf{Testing details}.
We resize input images with the longer side 1440 for IC13, 2000 for IC15,  1350 for Total-text and 1250 for CTW1500.  We set the number of fiducial points as 4 for two standard text datasets and 14 for two irregular text datasets. The detection results are given by connecting the predicted fiducial points.
Note that, all images are tested in the single-scale.

\subsection{Results on Standard Text Benchmarks}

\emph{Evaluation on horizontal text}.
We first evaluate our method on IC13 mainly consisting of horizontal texts.
Table \ref{tb:1} shows the results, and represents that our method achieve competitive performance compared to previous methods on the `Detection', `End-to-End' and `Word Spotting' evaluation items.
Besides, our method is also very efficient and achieves `10.3' of Frame Per Second (\emph{abbr.} FPS).

\emph{Evaluation on perspective text}.
We evaluate our method on IC15 containing many perspective texts, and the results are shown in Table \ref{tb:1}.
In the detection stage, our method achieves comparable performance with the irregular text spotting methods such as TextNet and Mask TextSpotter.
In the `End-to-End' and `Word Spotting' tasks, our method significantly outperforms previous irregular-text-based methods and achieves the remarkable state-of-the-art performance on general lexicon cases, which demonstrates the effectiveness of our method.

\subsection{Results on Irregular Text Benchmarks}

\begin{table}[ht]
\centering
\scalebox{0.78}{
\begin{tabular}{|l|c|c|c|c|c|}
\hline
\multirow{2}{*}{Method} & \multicolumn{3}{c|}{Detection} & \multicolumn{2}{c|}{End-to-End} \\ \cline{2-6}
 & P & R & F & None & Full\\
\hline
TextSnake \shortcite{long2018textsnake} & 82.7 & 74.5 & 78.4 & -& -\\
FTSN \shortcite{dai2018fused} & 84.7 & 78.0 & 81.3 & & \\
TextField \shortcite{xu2019textfield} & 81.2 & 79.9 & 80.6 & -& -\\
SPCNet \shortcite{xie2018scene} & 83.0 & 82.8 & 82.9 &- & -\\
CSE \shortcite{liu2019Towards} & 81.4 & 79.1 & 80.2 & - &- \\
PSENet-1s \shortcite{Wang2019Shape} & 84.0 & 78.0 & 80.9 & - & - \\
LOMO \shortcite{Zhang2019look} & 75.7 & \textbf{88.6} & 81.6 & - & -\\
Mask TextSpotter \shortcite{lyu2018mask} & 69.0 & 55.0 & 61.3 &  52.9  & 71.8\\
TextNet \shortcite{sun2018textnet} & 68.2 & 59.5 & 63.5 & 54.0 & - \\
\hline
Ours (2-stage) &  88.1 & 78.9 & 83.3 & 63.3 & 73.9 \\
Ours (End-to-end) &  \textbf{88.8} & 81.8 & \textbf{85.2} & \textbf{69.7} & \textbf{78.3} \\
\hline
\end{tabular}
}
\caption{Result on Total-Text. ``Full'' indicates lexicons of all images are combined. ``None'' means lexicon-free.}
\label{tb:3}
\end{table}

We test our method on two irregular text benchmarks: Total-Text and CTW1500, as shown in Table \ref{tb:3} and \ref{tb:4}.
In the detection stage, our method outperforms all previous methods and surpasses the best result 2.3\% on Total-Text and 2.4\% on CTW1500 on F-measure evaluation.

Moreover, our method significantly outperforms previous methods on the precision item, which attributes to the false-positive filtering strategy.
In the end-to-end case, our method significantly surpasses the best-reported results \cite{sun2018textnet} by 15.7\% on `None' and the best of results \cite{lyu2018mask} by 6.5\% on `Full', which mainly attributes to STM achieving the end-to-end training strategies.
Since CTW1500 releases the recognition annotation recently, there is no reported result on the end-to-end evaluation. Here, we report the end-to-end results lexicon-freely, and believe our method will significantly outperform previous methods.

\begin{table}[ht]
\centering
\scalebox{0.78}{
\begin{tabular}{|l|c|c|c|c|}
\hline
\multirow{2}{*}{Method} & \multicolumn{3}{c|}{Detection} & \multicolumn{1}{c|}{End-to-End} \\ \cline{2-5}
 & P & R & F & None \\
\hline
TextSnake \shortcite{long2018textsnake} & 69.7 & 85.3 & 75.6 & - \\
TextField \shortcite{xu2019textfield} & 83.0 & 79.8 & 81.4 & -\\
CSE \shortcite{liu2019Towards} & 81.1 & 76.0 & 78.4 & -  \\
PSENet-1s \shortcite{Wang2019Shape} & 84.8 & 79.7 & 82.2 & -\\
LOMO \shortcite{Zhang2019look} & 69.6 & \textbf{89.2} & 78.4 &  -\\
\hline
Ours (2-stage) & \textbf{88.7} & 78.2 & 83.1  & 48.6  \\
Ours (End-to-end) & 87.5 & 81.9 & \textbf{84.6}  & \textbf{57.0}  \\
\hline
\end{tabular}
}
\caption{Result on CTW1500. `None' means lexicon-free. }
\label{tb:4}
\end{table}

In summary, the results on Total-Text and CTW1500 demonstrate the effectiveness of our method for arbitrary-shaped text spotting.
Moreover, compared with 2-staged results, the end-to-end trainable strategy markedly boosts text spotting performance, especially for the recognition part.

\subsection{Ablation Results of Fiducial Points}
The number of fiducial points directly influences the detection and end-to-end results when texts are displayed in the curve or even waved shapes. Table \ref{ablation} shows the result that how the number of  fiducial points affects the detection and end-to-end evaluations on different benchmarks.
It is clear that 4 points annotation is enough for regular benchmark such as IC15, and there is almost no influence on the result when the number of fiducial points increases.
On the other hand, for two irregular benchmarks, the detection F-score as well as end-to-end F-score raises along with the increasing number of fiducial points, and the performance becomes stable when $2\times N\ge10$.
\begin{table}
\centering
\scalebox{0.78}{
\begin{tabular}{|l|c|c|c|c|c|c|c|c|}
\hline
\multirow{2}{*}{Dataset} & \multicolumn{8}{c|}{Number of fiducial points} \\ \cline{2-9}
 &  4 & 6 & 8 & 10 & 12 & 14 & 16 & 18 \\
\hline
IC15 & \textbf{87.1} & 87.0 & 87.0 & 86.9 & 87.0 & 86.9 & 86.8 & 86.8 \\
Total-Text & 71.5 & 82.8 & 84.5 & 85.0 & 85.2 & 85.2 & 85.2 & \textbf{85.3} \\
CTW1500 & 68.7 & 81.9 & 84.1 & 84.3 & 84.4 & \textbf{84.6} & 84.4 & 84.5 \\
\hline
\hline
Total-Text & 55.9 & 68.5 & 69.8 & 69.6 & 69.8 & 69.7 & 69.5 & \textbf{69.9} \\
CTW1500 & 40.2 & 52.2 & {56.2} & {57.0} & \textbf{57.1} & 57.0 & 56.5 & 56.4  \\
\hline
\end{tabular}
}
\caption{Detection (top part) and end-to-end (bottom part) evaluation (F-measure) under varied number of fiducial points for different benchmarks.}
\label{ablation}
\end{table}

\begin{figure}[t]
\centering
\includegraphics[width=\columnwidth]{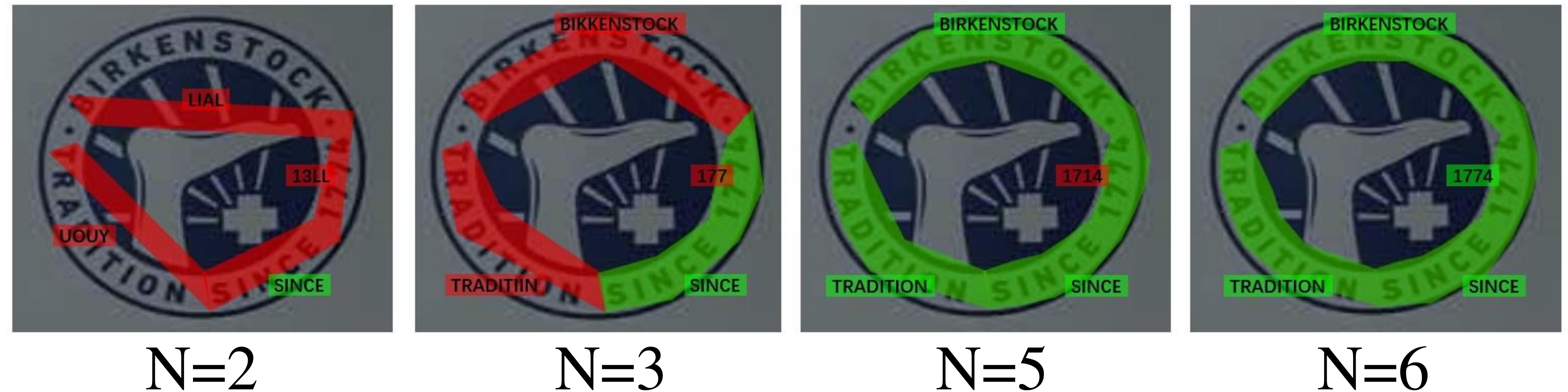}\\
\caption{Results of Text Perceptron with different number of fiducial points (4,6,10,12).}
\label{points}
\end{figure}

Figure \ref{points} shows an example of end-to-end evaluation under different number of fiducial points. We see that the generated text masks by few fiducial points are hard to cover the entire curve texts. As the growing number of fiducial points, STM has more power to catch and rectify irregular text instances, which yields higher recognition accuracy.

In contrast to previous works, our method can generate any fixed number of fiducial points on text boundaries. The fiducial points generation method  can also be used to annotate arbitrary-shaped text.

\subsection{Visualization Results}
\begin{figure}[ht]
	\centering
	\includegraphics[width=0.98\columnwidth]{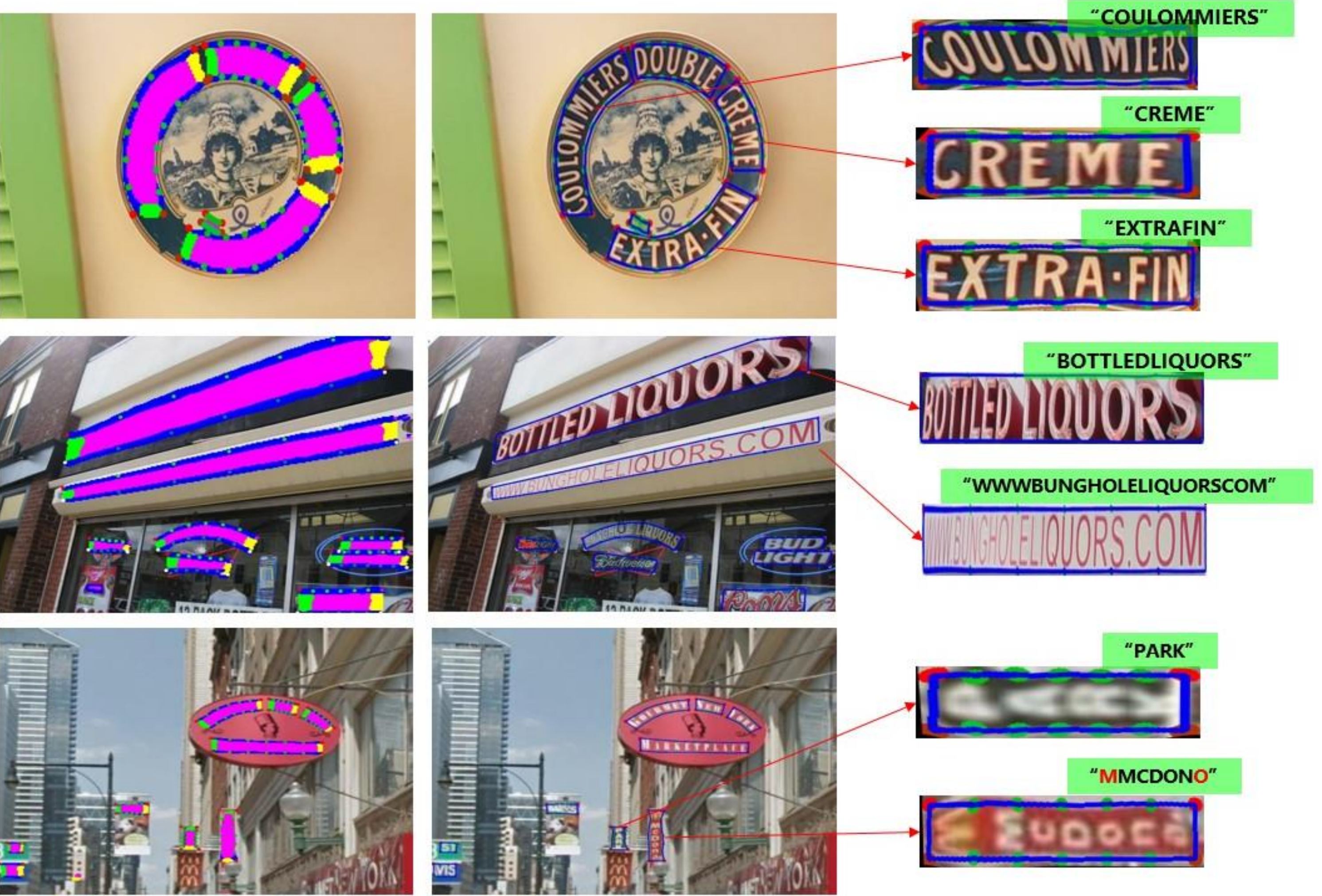}\\
	\caption{Visualization results on origin images.
	}
	\label{vis1}
\end{figure}
\begin{figure*}[htbp]
\begin{center}
\includegraphics[width=0.95\textwidth]{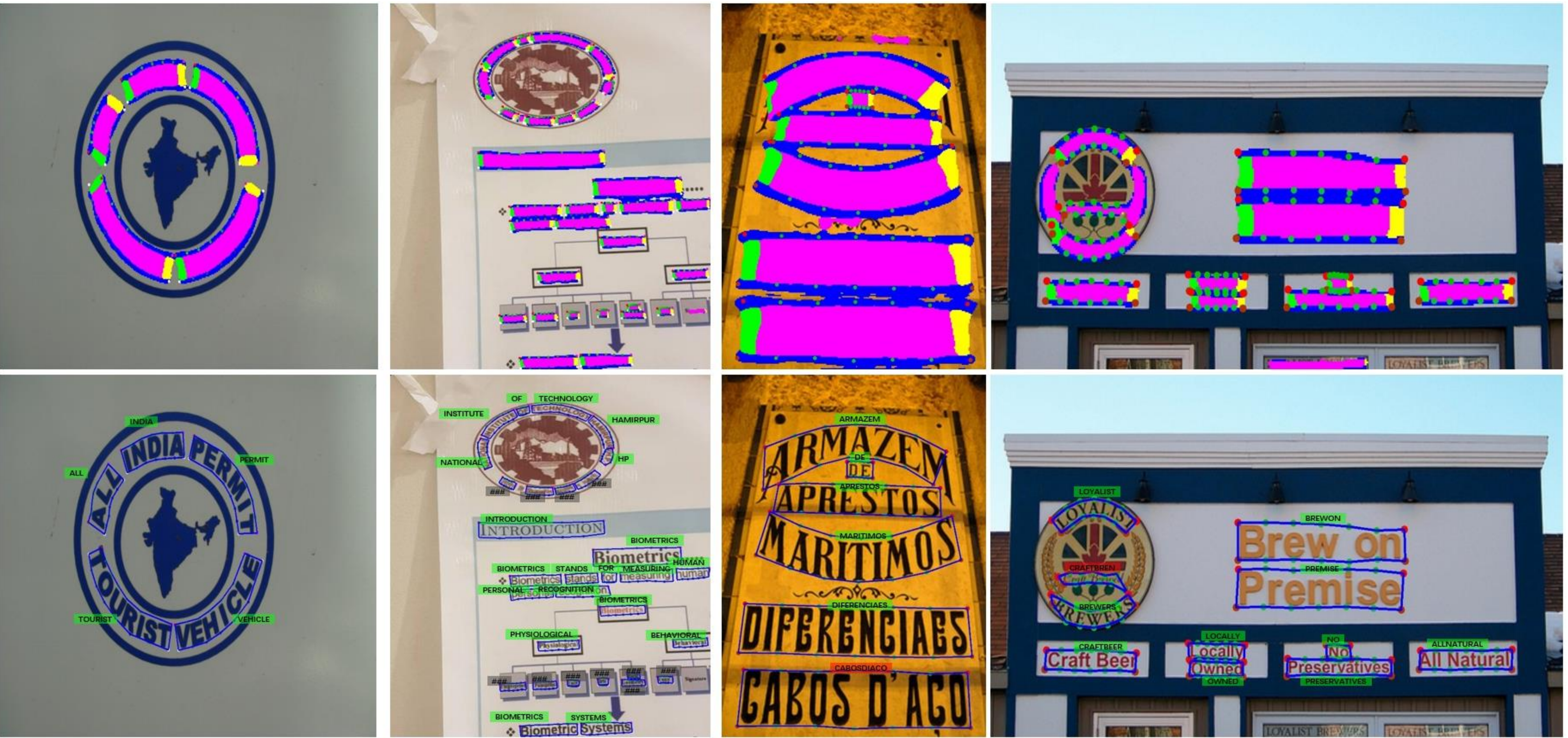}\\
\end{center}
   \caption{Visualization result on Total-Text and CTW1500. The first row displays the segmented results and the second row shows the end-to-end results. Fiducial points are also visualized as colored points on text boundaries.}
\label{vis2}
\end{figure*}

Figure \ref{vis1} and Figure \ref{vis2} demonstrate some visualization results in Total-Text and CTW1500 datasets. Text Perceptron shows its powerful ability in catching the reading order of irregular scene text  (including curved, long perspective, vertical, etc.), and with the help of fiducial points which can further recognize text in a much simpler way. From the segmentation results, we find that many of text-like false positives have been filtered out due to the missing of \emph{head} or \emph{tail} boundary. This means the features of \emph{head} or \emph{tail} boundaries contain the different semantic information with that of the \emph{center} region. Figure \ref{vis1} also shows the visualization of some rectified irregular text instances, in which vertical texts can be well transformed into the ``lying-down'' shapes.

\subsubsection{Failure Samples}
\begin{figure}[t]
\begin{center}
\includegraphics[width=0.8\columnwidth]{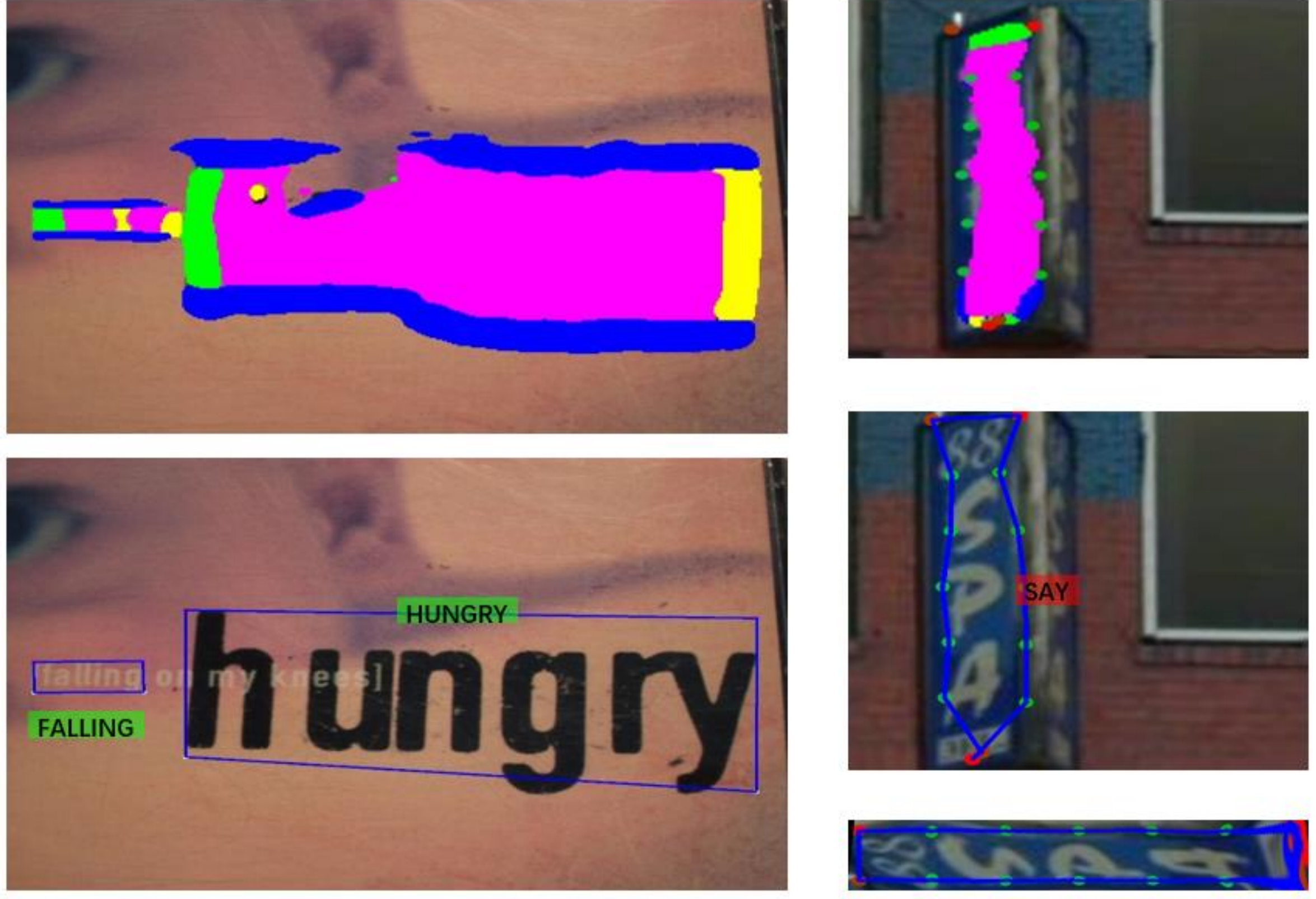}\\
\end{center}
   \caption{Visualization of some failure samples.}
\label{fail}
\end{figure}

We illustrate some failure samples that are difficult for Text Perceptron, as shown in Figure \ref{fail}.

\emph{Overlapped text}. It is a common tough task for segmentation-based detection methods. Pixels belong to the \emph{center} text region for one text instance may also become the boundary region for another one. Even though our orderly overlaying strategy allows pixels to have multiple classes and makes boundary pixels have higher priority than \emph{center} text pixels, which encourages inner instance to be separated from the outer instance. But experiments found that many times, the boundaries of inner instance cannot be fully recalled to embrace such instance, and connecting between \emph{center} text pixels will result in the failure of detecting such inner an instance.

\emph{Recognition of vertical instance}. On the one hand, vertical texts appear in little frequency in the common datasets. One the other hand, although Text Perceptron can read vertical instances from left to right, it is still a challenge for recognition algorithm to distinguish whether the instance is a horizontal text or a 'lying-down' vertical one. Therefore, there are some correctly detected instances cannot be recognized right. It is also a common difficult problem for all existing recognition algorithms.

\section{Conclusion}
In this paper, we propose an end-to-end trainable text spotter named {Text Perceptron} aiming at spotting text with arbitrary-shapes.
To achieve global optimization, a Shape Transform Module is proposed to unite the text detection and recognition into a whole framework. A segmentation-based detector is carefully designed to distinguish text instances and capture the latent information of text reading orders.
Extensive experiments show that our method achieves competitive result in standard text benchmarks and the state-of-the-art in both detection and end-to-end evaluations on popular irregular text benchmarks.

\newpage
\fontsize{8pt}{9pt} \selectfont
\bibliography{893_reference}
\bibliographystyle{aaai}

\end{document}